\pdfoutput=1

\documentclass[11pt]{article}

\usepackage{acl}

\usepackage{times}
\usepackage{latexsym}

\usepackage[T1]{fontenc}

\usepackage[utf8]{inputenc}

\usepackage{microtype}

\usepackage{inconsolata}

\usepackage{lipsum}
\usepackage{soul}
\usepackage{graphicx}
\usepackage{booktabs}
\usepackage{cleveref}
\usepackage{csquotes}

\usepackage{listings}
\usepackage[strict]{changepage}
\usepackage{framed}

\definecolor{formalshade}{rgb}{0.95,0.95,1}
\definecolor{promptshade}{rgb}{0.95,0.95,0.5}

\newenvironment{formal}{%
  \MakeFramed{\advance\hsize-\width\FrameRestore}%
  \noindent\hspace{-4.55pt}%
  \begin{adjustwidth}{}{7pt}%
  \vspace{-1pt}\vspace{0pt}%
}
{%
  \vspace{2pt}\end{adjustwidth}\endMakeFramed%
}

\newenvironment{prompt}{%
  \MakeFramed{\advance\hsize-\width\FrameRestore}%
  \noindent\hspace{-4.55pt}%
  \begin{adjustwidth}{}{7pt}%
  \vspace{-1pt}\vspace{0pt}%
}
{%
  \vspace{2pt}\end{adjustwidth}\endMakeFramed%
}

\title{Native Language Identification with Large Language Models}

\author{Wei Zhang\textsuperscript{1} \and Alexandre Salle\textsuperscript{2}\\
    \textsuperscript{1} Miami, Florida, USA \\
    \textsuperscript{2} VTEX, Porto Alegre, BR \\
    \texttt{alex@alexsalle.com}
}

\begin{document}
\maketitle
\begin{abstract}

We present the first experiments on Native Language Identification (NLI) using LLMs such as GPT-4. NLI is the task of predicting a writer's first language by analyzing their writings in a second language, and is used in second language acquisition and forensic linguistics.
Our results show that GPT models are proficient at NLI classification, with GPT-4 setting a new performance record of 91.7\% on the benchmark TOEFL11 test set in a zero-shot setting.
We also show that unlike previous fully-supervised settings, LLMs can perform NLI without being limited to a set of known classes, which has practical implications for real-world applications.
Finally, we also show that LLMs can provide justification for their choices, providing reasoning based on spelling errors, syntactic patterns, and usage of directly translated linguistic patterns.
\end{abstract}

\section{Introduction}

\begin{figure}
    \centering
    \includegraphics[width=\linewidth]{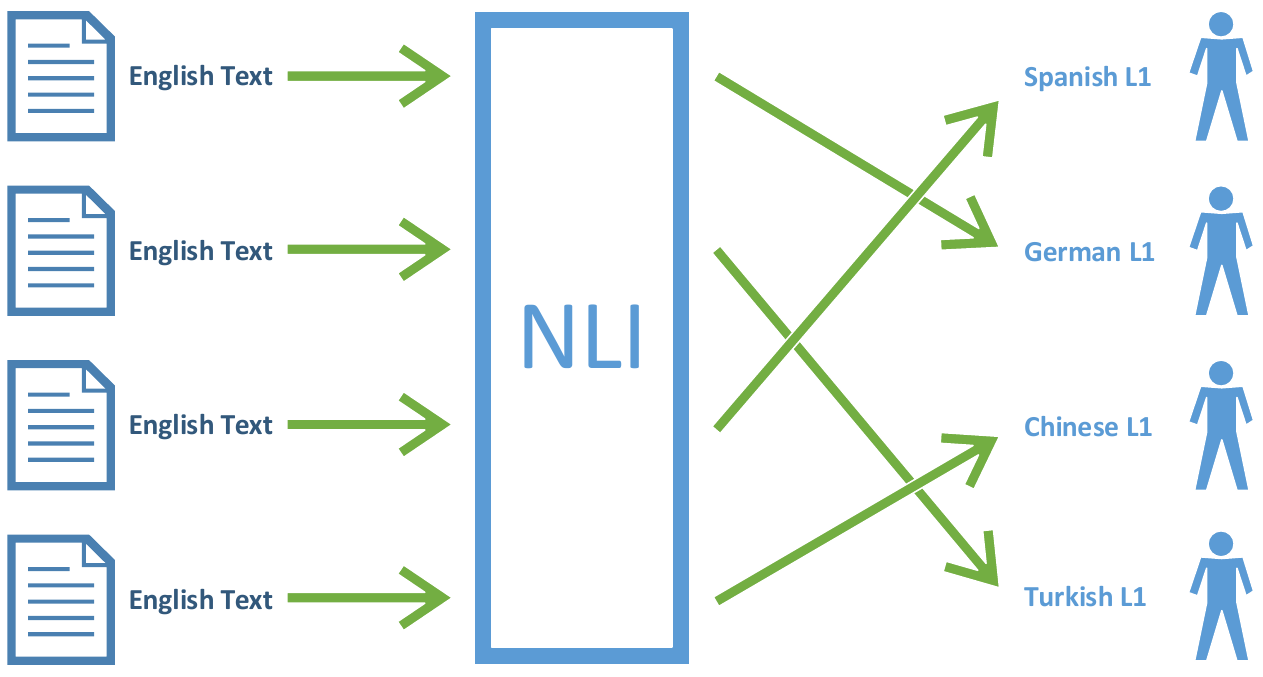}
    \caption{A conceptualization of NLI: Texts in English penned by non-native speaker authors are processed by an NLI system that predicts the author's mother tongue (L1) through the analysis of their writing style. Reproduced from \citet[p. 21]{malmasi2016native}}
    \label{fig:nli}
\end{figure}

Native Language Identification (NLI) is the task of identifying an individual's native language based on their writing or speech.
Text-based NLI involves determining the first language (L1) of an author by examining a text they have written in a second language (L2), as shown in \Cref{fig:nli}.
NLI detects patterns in language usage that are characteristic of individuals sharing the same native language. The underlying assumption is that a writer's native language predisposes them to exhibit specific patterns in their second language, influenced by their mother tongue's linguistic structure.

The task has gained attention within the Natural Language Processing sub-field due to its diverse applications. NLI can be used in language learning and teaching, where it can help in identifying a learner's native language and customize the learning material accordingly. In forensic linguistics, NLI can be an investigative tool to guess the native language of an unknown author.

Traditionally, NLI is framed as a supervised learning task where a model is trained on second language (L2) text that is annotated with real first language (L1) labels. This method relies on a model's ability to learn and generalize from the labeled training data in order to accurately predict the native language of new, unlabeled texts.

Until recently, traditional Machine Learning approaches based on hand-crafted linguistic features outperformed Deep Learning methods for NLI \cite[p. 72]{malmasi-etal-2017-report}.
More recently, larger Transformer architectures like GPT-2 have been applied for the task and improved on those results.

However, while newer Large Language Models (LLMs) like GPT-4 have set new state-of-the-art benchmarks across various NLP tasks \cite{openai2023gpt4}, their application in NLI remains unexplored. Given their advanced understanding capabilities, LLMs hold the promise of significantly enhancing NLI performance by discerning nuanced language patterns indicative of native language influence.

Accordingly, we aim to apply LLMs for NLI, with the expectation that their sophisticated understanding of linguistic nuances will offer improvements in NLI performance and explainability.

Our contributions are as follows:

\begin{itemize}
    \item We present the first NLI experiments using GPT-3.5 and GPT-4.
    \item We show that GPT-4 set a new performance record of 91.7\% on the TOEFL11 benchmark.
    \item We demonstrate that zero-shot NLI can be performed without specifying the list of known L1 classes.
    \item We examine the capability of GPT-4 in providing linguistic reasoning for its decisions.
\end{itemize}

\section{Related Work}

Previous NLI work has successfully used traditional ML models (such as Logistic Regression or SVMs) trained on extensive automatically extracted linguistic features \cite{malmasi-cahill-2015-measuring}, with these models performing well by learning  L1 patterns appearing in L2 texts.
An extensive examination of these approaches is presented by \citet{malmasi2016native}.
These approaches were found to outperform Deep Learning methods at the time \cite{malmasi-etal-2017-report}, most likely because early embedding-based approaches were limited in parameter size (relative to current multi-billion parameter models) and were able to capture semantics much more than the stylistic cues critical for NLI.

More recently, Transformer-based models, which represent the current cutting-edge in NLP, have been applied to NLI, potentially offering more sophisticated pattern recognition and an improved understanding of language nuances related to L1 influence.
\citet{steinbakken-gamback-2020-native} applied BERT to different NLI datasets, but were unable to surpass existing SVM ensemble approaches such as those of \citet{malmasi-dras-2018-native} which achieved an accuracy of 86.8\% on TOEFL11.
\citet{lotfi-etal-2020-deep} applied GPT-2 to NLI by fine-tuning one model per L1. At inference time, an L2 text is run through each model and the LM loss is computed, and the L1 label of the model with the lowest loss is assigned. This approach established a new SOTA result of 89\% accuracy on the TOEFL11 test set.
A key shortcoming of this approach is the expensive training and inference process (not to mention the memory and storage requirements).
\citet{uluslu-schneider-2022-scaling} attempted to remedy this by integrating Transformer Adapters into GPT-2 to enable efficient parameter sharing across all the L1 models. While this approach improved inference speed by 13x, its performance dropped to 84.2\% which underperforms the SVM ensemble approach.

Our experiments build on this prior work, but differ in key aspects. Firstly, we apply much larger models such as GPT-4 to the task for the first time. Secondly, we perform the task in a zero-shot setting and do not leverage any NLI-specific data. Finally, we explore new directions enabled by the emerging capabilities of the latest LLMs: open-set classification to move NLI beyond the supervised paradigm, as well as leveraging the LLMs to extract explainable discriminative features.

\section{Method and Data}

Our experiments are conducted in a zero-shot setting, evaluating models without the use of any training data that is specific to the NLI task. Instead, the models rely solely on their pre-existing knowledge and understanding gained from standard pretraining, to infer the native language of a text's author. 

\subsection{Models}
We experiment with the following LLMs, which are currently considered the top-performing models.

\paragraph{GPT-3.5}
This is an updated version of the GPT-3 model \cite{brown2020language} which has enhanced instruction-following capabilities and is trained using RLHF. 

\paragraph{GPT-4}
The latest version in the GPT model family, and widely considered to be the most capable LLM created to date. It has been shown to outperform GPT-3.5 for many tasks.

Our work leverages the \texttt{gpt-3.5-turbo} and \texttt{gpt-4} models respectively, and both were accessed via the OpenAI API.

\subsection{Dataset}
We utilize the TOEFL11 dataset\footnote{\url{https://catalog.ldc.upenn.edu/LDC2014T06}} for our experiments, which is the \textit{de facto} benchmark for NLI research.
Also known as the 
ETS Corpus of Non-Native Written English \cite{blanchard2013toefl11}, 
the dataset comprises 1,100 English essays by native speakers of 11 diverse languages: Arabic (ARA), Chinese (CHI), French (FRE), German (GER), Hindi (HIN), Italian (ITA), Japanese (JPN), Korean (KOR), Spanish (SPA), Telugu (TEL), and Turkish (TUR). There are 12,100 essays in total, with an average length of 348 words. These essays are responses to eight distinct prompts presented to all 11 L1 language groups and are authored by individuals with varying levels of English proficiency (low, medium, and high).
Since we work in a zero-shot setting, we do not leverage the entire TOEFL11 corpus, but only the test set which contains 1,100 essays (100 per L1).

\paragraph{Data Leakage}
While the dataset is available to researchers, to the best of our knowledge the texts are not publicly available, making it probable that the LLMs we use have not been pretrained on this data. This allows the test set to be evaluated in a zero-shot setting. Given this setup, we only employ the test set from TOEFL11 dataset, assessing the LLMs' ability to identify the L1 without any prior task-specific fine-tuning, to genuinely assess their zero-shot learning capabilities.

\subsection{Evaluation Metrics}
In line with prior work, we employ accuracy as the primary evaluation metric.
This is appropriate since our test set is balanced, ensuring that accuracy reflects true model performance without bias. We benchmark our results against those reported in recent papers, and we also include a comparison against a naive random guess baseline.

\section{NLI Classification Experiment}
\label{sec:exp1}

In our first experiment, we replicate traditional NLI classification using LLMs. This is a closed-set task, meaning that the model's predictions are limited to the predefined set of 11 L1 classes corresponding to the different native languages. We aim to directly compare the performance of LLMs against the existing state-of-the-art methods.

Each document from the TOEFL11 test set is input to the LLM for prediction.
The exact prompts we use are listed in \Cref{app:prompts}.
If the model predicts an L1 outside the set of known L1 classes, we apply iterative prompting and provide feedback that the response is incorrect, asking the model to refine its classification. This process is repeated until one of the known classes is predicted.

\begin{table*}[!hbt]
    \centering
    \begin{tabular}{|l|c|}
    \hline
    \textbf{Model} & \textbf{TOEFL11 Test Set} \\
    \hline
    Random Guess Baseline  & 9.1\%\\
    \hline
    SVM + Meta-Classifier \cite{malmasi-dras-2018-native} & 86.8\% \\
    BERT + Meta-Classifier \cite{steinbakken-gamback-2020-native} & 85.3\% \\
    GPT-2 \cite{lotfi-etal-2020-deep} & 89.0\% \\
    \hline
    Ours - GPT-3.5 (Zero-shot) & 74.0\% \\
    Ours - GPT-4 (Zero-shot) & \textbf{91.7\%} \\
    \hline
    Ours - GPT-3.5 (Open-set, Zero-shot) & 73.4\% \\
    Ours - GPT-4 (Open-set, Zero-shot)  & 86.7\% \\
    \hline
    \end{tabular}
    \caption{Results for our method on the TOEFL11 test set. We report accuracy, and prior SOTA baselines are listed.}
    \label{tab:results}
\end{table*}

We compare the performance of GPT-3.5 and GPT-4, evaluating them against three previous models. All results are shown in \Cref{tab:results}.
The GPT models perform well in a zero-shot setting. GPT-4 outperforms all previous models with an accuracy of 91.7\%, a +2.7\% increase over the previous best result and a new state-of-the-art for TOEFL11.
This is an impressive result given that no training or feature engineering was performed.
In fact, the GPT-4 top-1 accuracy approaches the top-2 accuracy of 92.2\% on this test set based on 29 distinct NLI systems (which is an easier setting) \cite[Table 3]{malmasi-etal-2015-oracle}.
This performance also suggests that LLM-based NLI is far more generalizable than previous work, where cross-corpus studies have shown large performance drops when testing NLI on out-of-distribution data \cite{malmasi-dras-2015-large}.

\begin{figure}
    \centering
    \includegraphics[width=\linewidth, clip=True,trim=0 0.32cm 0 0.2cm]{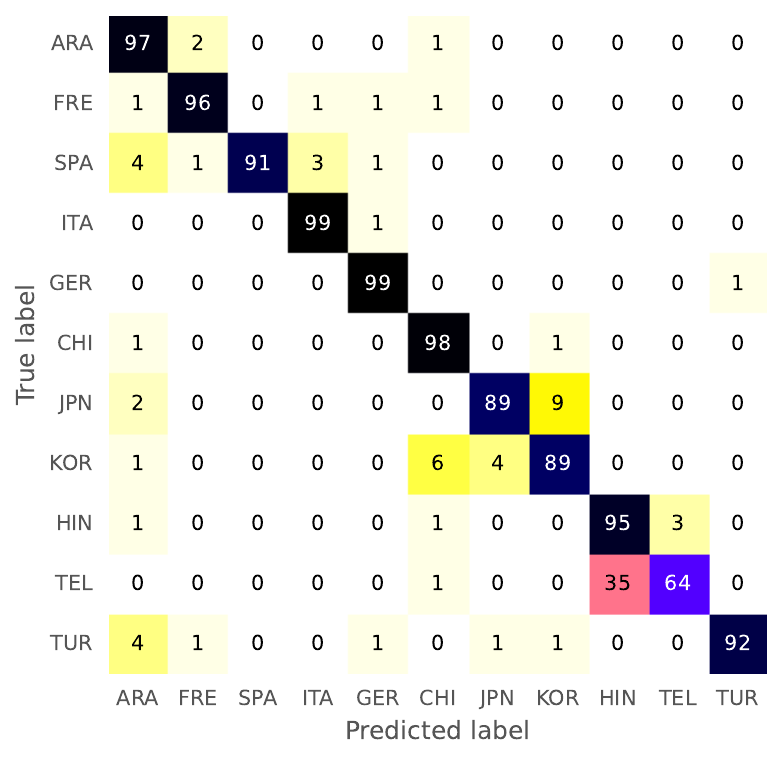}
    \caption{Confusion Matrix for the TOEFL11 test set for the predictions from GPT-4.}
    \label{fig:cm-gpt-4}
\end{figure}

A confusion matrix of the GPT-4 predictions (\Cref{fig:cm-gpt-4}) reveals that the greatest confusion is between Hindi and Telugu L1s. This is a well-known difficulty, and since both languages are from India, it is hypothesized to be due to the variety of English taught in the country \cite[p. 131]{malmasi-etal-2013-nli}.
Another cluster of errors occurs between Chinese, Japanese, and Korean.

GPT-3.5 only achieves an accuracy of 74\%.
The poor performance of GPT-3.5 can be attributed to the fact that for 12\% of the test set the L1 is initially predicted as English, and when asked to refine this classification, the model almost always chooses French as the label.
This is likely a result of the smaller GPT model's inability to identify distinguishing features in the text. We will further explore this aspect in \Cref{sec:open-set-results}.

\subsection{Impact of Text Length}
We also assess the impact of L2 text length on LLM-based NLI performance.
This experiment involves feeding the language models with segments of text at varying lengths, up to the first 2,000 characters of each document.\footnote{The TOEFL11 test documents have a mean length of 1,785 characters (SD=$\pm$459), and we chose this instead of tokens for simplicity. Documents may be shorter than the maximum input length, and such documents are included.}
This assessment helps us understand the relationship between text length and the ability of the models to accurately determine the writer's native language.

\begin{figure}
    \centering
    \includegraphics[width=\linewidth]{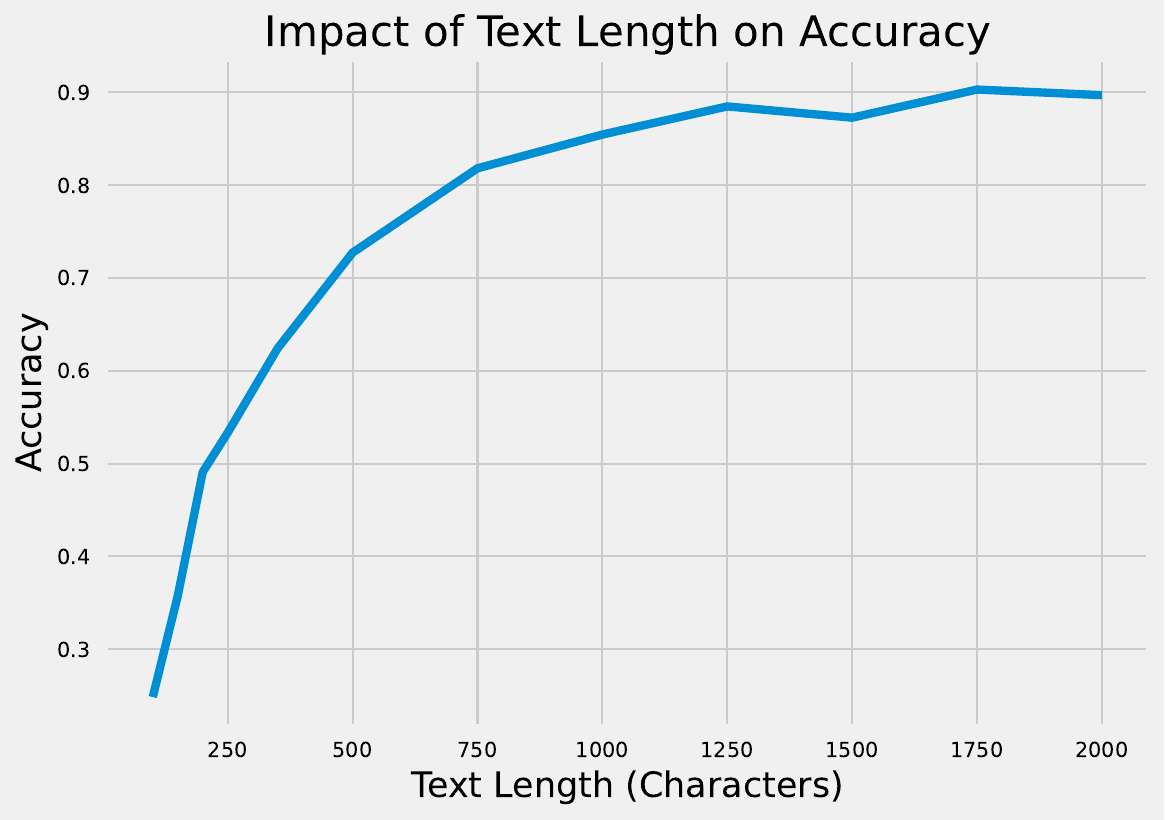}
    \caption{NLI performance (accuracy) for GPT-4 when using different input text lengths.}
    \label{fig:length-curve}
\end{figure}

We used GPT-4 for this experiment, and performance (accuracy) for different input lengths is shown in \Cref{fig:length-curve}. This result demonstrates that performance improves with text length, plateauing at around 1,250 characters for the TOEFL11 test set. This finding has implications for real-world NLI applications, indicating that a minimum amount of text may be required to achieve reliable results.\footnote{For inputs less than 100 characters, the GPT models often refused to make a classification due to insufficient evidence.}

\section{Open-Set Classification Experiment}
\label{sec:open-set-results}

Since we work in a zero-shot setting, our set of output classes does not need to be limited to the known set of classes in the dataset.
In fact, this has been a persistent shortcoming of previous NLI work as the supervised models used in most studies may not be practical for real-world applications since they are limited to predicting L1s contained in the training set.
Our second experiment aims to assess NLI classification performance in this setting, where the task is performed with a simplified instruction prompt. We also analyze the out-of-set L1 classes predicted by each model.

We do this by removing the output class restrictions from our prompt, thus allowing the model to output any desired language label (see \Cref{app:prompts} for prompts). Post-processing using regular expressions based on the output format specified in the prompt is then applied to parse the model output and extract the predicted language, and finally map it to a known class label. If the extracted output language cannot be mapped to a known class, a random class label is assigned to that test document in order to perform evaluation on the entire test.\footnote{Ideally the closest language in the known set (as defined by some linguistic metric) should be assigned, but we leave this for future work as maximizing classification accuracy is not the primary aim of this experiment.} 

Classification results are shown in the bottom section of \Cref{tab:results}.
We see that performance drops for both GPT models compared to the closed-class setting, with a greater relative drop for GPT-4.
To better understand this, we need to look at the set of out-of-set classes predicted by each model. The complete set of unknown L1 classes and the counts of their actual L1 labels are shown in \Cref{tab:oos-gpt35} for GPT-3.5 and \Cref{tab:oos-gpt4} for GPT-4.

\begin{table*}[ht]
    \center
    \resizebox{\textwidth}{!}{
    \begin{tabular}{lccccccccccc|r}
\toprule
GPT-3.5 Predicted L1 & ARA & CHI & FRE & GER & HIN & ITA & JPN & KOR & SPA & TEL & TUR & Total \\
\midrule
English & 6 & 2 & 1 & 2 & 53 & 1 & 2 & 3 & 4 & 44 & 8 & 126 \\
Tamil & 0 & 0 & 0 & 0 & 0 & 0 & 0 & 0 & 0 & 12 & 1 & 13 \\
Portuguese & 0 & 0 & 0 & 0 & 1 & 0 & 0 & 0 & 3 & 0 & 1 & 5 \\
Bengali & 0 & 0 & 0 & 0 & 0 & 0 & 0 & 0 & 0 & 3 & 0 & 3 \\
Persian & 0 & 0 & 0 & 0 & 0 & 0 & 0 & 0 & 0 & 0 & 2 & 2 \\
Dutch & 0 & 0 & 1 & 0 & 0 & 0 & 0 & 0 & 0 & 0 & 0 & 1 \\
Indeterminable & 0 & 0 & 0 & 0 & 1 & 0 & 0 & 0 & 0 & 0 & 0 & 1 \\
Malay & 0 & 1 & 0 & 0 & 0 & 0 & 0 & 0 & 0 & 0 & 0 & 1 \\
Vietnamese & 0 & 0 & 0 & 0 & 0 & 0 & 0 & 1 & 0 & 0 & 0 & 1 \\
\bottomrule
\end{tabular}
}
\caption{Counts of out-of-set L1 classes (rows) predicted by GPT-3.5 on the TOEFL11 test set. Counts of the true L1 classes are given in the columns.}
\label{tab:oos-gpt35}
\end{table*}

As shown in \Cref{tab:oos-gpt35}, GPT-3.5 predicts 126 of the instances as English, with the majority of those actually belonging to the Hindi and Telugu classes. Unlike our first experiment, we do not prevent the model from predicting English as the L1, and this failure to recognize non-native text is the largest source of the performance gap between GPT-3.5 and GPT-4.
The other predicted classes are all related to the actual L1 class either linguistically or geographically.
On the other hand, GPT-4 does not predict any of the documents as an English L1 text, as seen in \Cref{tab:oos-gpt4}. Some of the out-of-set predicted L1 classes are linguistically or geographically close to the ground truth labels, but several other are not.

\begin{table*}[ht]
\center
\begin{tabular}{lcccccccc|r}
\toprule
GPT-4 Predicted L1 & CHI & FRE & HIN & ITA & KOR & SPA & TEL & TUR & Total \\
\midrule
Russian & 0 & 1 & 0 & 0 & 1 & 0 & 0 & 5 & 7 \\
Persian (Farsi) & 0 & 0 & 0 & 0 & 0 & 0 & 0 & 4 & 4 \\
Dutch & 0 & 0 & 0 & 0 & 1 & 1 & 1 & 0 & 3 \\
Indian Language & 0 & 0 & 0 & 0 & 0 & 0 & 2 & 0 & 2 \\
Amharic & 0 & 0 & 0 & 0 & 1 & 0 & 0 & 0 & 1 \\
Bengali & 0 & 0 & 1 & 0 & 0 & 0 & 0 & 0 & 1 \\
Malay (Malaysian) & 1 & 0 & 0 & 0 & 0 & 0 & 0 & 0 & 1 \\
Portuguese & 0 & 0 & 0 & 0 & 0 & 1 & 0 & 0 & 1 \\
Romanian & 0 & 0 & 0 & 1 & 0 & 0 & 0 & 0 & 1 \\
Tamil & 0 & 0 & 1 & 0 & 0 & 0 & 0 & 0 & 1 \\
\bottomrule
\end{tabular}
\caption{Counts of out-of-set L1 classes (rows) predicted by GPT-4 on the TOEFL11 test set. Counts of the true L1 classes are given in the columns.}
\label{tab:oos-gpt4}
\end{table*}

This result demonstrates that GPT models are able to perform open-set NLI, which has practical implications for real-world applications.

\section{Discriminative Feature Explainability}
\label{sec:explainability}

NLI classification can be valuable in forensic linguistics and investigative scenarios, aiding experts in deciphering textual evidence.
Conversely, in language learning contexts, the focus shifts from mere classification accuracy to the identification and interpretation of linguistic features that distinguish L1s. This underscores NLI models' utility for data exploration and providing explanations, offering insights into the specific language characteristics writers exhibit, which can inform tailored language instruction and support linguistic research.

\citet{malmasi-dras-2014-language} were the first to look at this in the context of NLI, using the weights of an SVM model to identify potential language transfer hypotheses (L2 usage patterns that are transferred from the writer's L1). Others have used similar approaches to identify native-like expressions \cite{harust-etal-2020-native}, or applied Explainable Machine Learning to NLI \cite{10.1093/llc/fqad019}.

Our final experiment investigates whether LLMs can offer explanations for their NLI predictions. Previous approaches have relied heavily on feature engineering and the creation of feature representations to interpret models' decision-making processes \cite{malmasi-dras-2014-language}. In contrast, our approach seeks to leverage LLMs in a zero-shot fashion to generate natural language explanations and provide cited examples directly. This would not only streamline the process of understanding the linguistic cues used for NLI but could also offer immediate, comprehensible insights into the model's reasoning, which is especially beneficial for language learning and forensic analysis.

We adapt the prompt used in previous experiments, and ask the model to provide reasoning and examples to justify the chosen class label (see \Cref{app:prompts} for the exact prompt). We then parse the model output to extract the explanations, resulting in responses such as:

\begin{formal}
The text is quite well-written, with only minor indicators pointing towards the author's native language being German. The major clue is the author's use of reflexive verbs in instances where they are not commonly used by native English speakers. For example, ``the consumer's responsibility to train itself'' - a native English speaker would typically phrase this as ``it's the consumer's responsibility to train themselves''.

This is a reflection of the regular use of reflexive verbs in German. In addition, the phrase, ``they are fake, so why do we believe them anyway?'' is syntax reminiscent of German structure. The use of colons before explanations, though not incorrect in English, also suggest German as it is more prevalent in German writing, such as in ``realized: the advertisement''.
\end{formal}

Such justifications were obtained for the complete test set using GPT-4.
While a detailed analysis of the explanations is outside the scope of the current work, an examination of the reasoning revealed many reasonable or seemingly viable claims.
This outcome indicates that sufficiently-large LLMs can be used as tools for linguistic analysis of learner language. However, while they are able to justify their predictions, it is not clear if they can be used to develop novel language transfer hypotheses. That is a process which requires examining large numbers of L2 texts to find common linguistic patterns that can be linked to the shared L1. 
In the next section we provide some examples of the common linguistic features cited by the GPT models in their reasoning.

\subsection{Common Linguistic Features}

\paragraph{Spelling and Typography}
The model picks up on peculiar spelling errors and attempts to interpret them within the context of the selected L1. Two examples from German and Arabic L1 texts are shown below:

\begin{formal}
The author spells "proven" as "prooven," "interest" as "interrest," and "horizon" as "horrizon." These spelling errors indicate an attempt to phonetically recreate words, which may suggest German as their native language since double vowels in German often alter the pronunciation of a word (e.g., "Boot" for boat).
\end{formal}

\begin{formal}
There are spelling errors like "descuss" instead of "discuss" and "medecin" instead of "medicine" which could suggest a difficulty with phonetics which is common in Arabic speakers as English contains some phonemes that do not exist in Arabic.
\end{formal}

\paragraph{Syntax and Grammar}
Various types of syntactic patterns are frequently pointed out by the model.

Spanish examples:

\begin{formal}
Preposition misuse: ``related with reality'' should be ``related to reality''. This mistake is commonly made by Spanish speakers because the prepositions `to' and `with' translate to a single word, `con', in Spanish.
\end{formal}

\begin{formal}
Misuse of infinitive forms: In Spanish, infinitives often stand alone as nouns. Here, the author exhibits this sort of error by using a bare infinitive after "of" in the phrase "advantages of to make analogies".
\end{formal}

Chinese example:

\begin{formal}
Omission of articles: In Chinese, there are no articles ('a', 'an', 'the'). This lack of articles leads to errors in English such as ``Understanding more knowledge could serve as path for people.''
\end{formal}

\paragraph{Translation and Transliteration}
GPT-4 sometimes links a peculiar phrase to a common structure in the L1, and identifies direct translation or transliteration on the part of the author as the root cause.

Italian:

\begin{formal}
Another common mistake for Italian speakers is confusing "is" and "it's". In Italian, verbs aren't articulated with auxiliary verbs as in English. For instance, the text writes, "Is important understand," which could be a direct translation from the Italian "È importante capire," where "È" translates to "is".
\end{formal}

\pagebreak

Spanish:

\begin{formal}
In the sentence "the ones who focus their energies on a particular subject are being on the risk of becoming too good on the preferred bussiness", the author uses "being on the risk" which seems like a direct translation from Spanish "estar en riesgo" instead of the more common English phrase "being at risk".
\end{formal}

\subsection{Accuracy and Hallucinations}

Manual examination of the explanations revealed a number of hallucinations and incorrect assertions. However, correctly evaluating the accuracy of the explanations requires an expert human study, and is beyond the scope of this work.
Some examples of hallucinated explanations are shown below:

\begin{formal}
Second, the writer swaps "v" and "f" sounds ("affort" instead of "effort"), two sounds that are not distinguished in many Arabic dialects.
\end{formal}

\begin{formal}
Misplaced adjectives: The phrase `my fist job'. In Spanish, adjectives are typically placed after the noun they modify, which might lead Spanish speakers to sometimes misplaced the adjectives when writing English. In this case, the Spanish equivalent would be `mi primer trabajo', which might have made the author write `fist' instead of `first', thus indicating that their mother tongue is Spanish.
\end{formal}

\section{Conclusion}
We presented the first NLI experiments using LLMs such as GPT-4.
Our results show that GPT models are adept at performing NLI classification, with GPT-4 setting a new SOTA performance of 91.7\% on the benchmark TOEFL11 test set.

Our second experiment demonstrated that LLM-based NLI can be performed in an open-set manner, without specifying the list of known L1 classes.
Finally, we also showed that LLMs hold great promise for explainable L2 data analysis, potentially enabling educators and linguists to unravel the nuances of language use that differentiate native speakers from learners. This explanatory capacity can thus inform tailored language instruction and contribute to a deeper understanding of language acquisition processes.

\section*{Limitations and Future Work}

\paragraph{Improved Prompting}
While we prompted the model to provide reasoning for its choice, it's likely possible to obtain even more detailed analysis with a more specific prompt.
Further, it is also possible to explore the generation of counter-explanations or alternative hypotheses.

\paragraph{Open-Source LLMs}
Additionally, fine-tuning of open-source LLMs (such as Llama-2) for NLI remains unexplored. While we do not expect such models to outperform GPT-4 (based on other NLP benchmarks), it may be worthwhile to investigate their performance and understand the gaps between these models.

\paragraph{Multilingual Evaluation}
NLI has been applied to many languages \cite{malmasi2017multilingual}, including Arabic \cite{malmasi-dras-2014-arabic} and Chinese \cite{malmasi-dras-2014-chinese}. Our work only evaluated English L2 data as this is the benchmark dataset for the task. Future experiments can assess multilingual variants of the task.

\bibliography{anthology,custom}

\appendix
\onecolumn

\section*{\center Appendix}

\section{GPT Prompts}
\label{app:prompts}

For the standard (closed-set) classification task we use the following System prompt:

\begin{prompt}
\begin{lstlisting}
You are a forensic linguistics expert that reads English texts written by non-native authors in order to classify the native language of the author as one of:

"ARA": Arabic
"CHI": Chinese
"FRE": French
"GER": German
"HIN": Hindi
"ITA": Italian
"JPN": Japanese
"KOR": Korean
"SPA": Spanish
"TEL": Telugu
"TUR": Turkish

Use clues such as spelling errors, word choice, syntactic patterns, and grammatical errors to decide. 

DO NOT USE ANY OTHER CLASS.
IMPORTANT: Do not classify any input as "ENG" (English). English is an invalid choice.

Valid output formats:
Class: "ARA"
Class: "CHI"
Class: "FRE"
Class: "GER"
\end{lstlisting}
\end{prompt}

The input text is then provided as an input (user) prompt:

\begin{prompt}
\begin{lstlisting}
<TOEFL11 ESSAY TEXT>

Classify the text as one of ARA, CHI, FRE, GER, HIN, ITA, JPN, KOR, SPA, TEL, or TUR. Do not output any other class - do NOT choose "ENG" (English). What is the closest native language of the author of this English text from the given list?
\end{lstlisting}
\end{prompt}

For the open-set task we use the following System prompt:

\begin{prompt}
\begin{lstlisting}
You are a forensic linguistics expert that reads texts written by non-native authors in order to identify their native language.

Analyze each text and identify the native language of the author.

Use clues such as spelling errors, word choice, syntactic patterns, and grammatical errors to decide.
\end{lstlisting}
\end{prompt}

The input text is then simply passed as a user prompt.

For the explainability experiment, we request the following output in the System prompt:

\begin{prompt}
\begin{lstlisting}
You must provide a guess. Output two named sections: (1) "Native Language" with the name of the language, and (2) "Reasoning" with a detailed explanation of your judgement with examples from the text.
\end{lstlisting}
\end{prompt}
\end{document}